\documentclass{article}

    \usepackage[final]{neurips_2024}

\usepackage[utf8]{inputenc} 
\usepackage[T1]{fontenc}    
\usepackage{hyperref}       
\usepackage{url}            
\usepackage{booktabs}       
\usepackage{amsfonts}       
\usepackage{nicefrac}       
\usepackage{microtype}      
\usepackage{xcolor}         
\usepackage{graphicx}
\usepackage{amsmath}
\usepackage[longend]{algorithm2e}
\usepackage{tcolorbox}
\usepackage{multirow}
\usepackage{array}
\usepackage{arydshln}
\usepackage{changepage}
\usepackage{multicol}
\usepackage{enumitem}

\title{Words as Beacons: Guiding RL Agents with High-Level Language Prompts}

\author{%
  Unai Ruiz-Gonzalez$^{1,2}$\\
  \texttt{ruizgonzalezunai@gmail.com} \\
  \And
  Alain Andres$^{1}$ \\
  \texttt{alain.andres@tecnalia.com} \\
  \And
  Pedro G. Bascoy$^1$ \\
  \texttt{pedro.gonzalez@tecnalia.com} \\
  \And
  Javier Del Ser$^{1,2}$ \\
  \texttt{javier.delser@tecnalia.com} \\
  \AND
  \textnormal{$^1$ TECNALIA, Basque Research and Technology Alliance (BRTA), Donostia-San Sebastian, Spain} \\
  \textnormal{$^2$ University of the Basque Country (UPV/EHU), Bilbao, Spain} \\
}

\begin{document}
\maketitle

\begin{abstract}
  Sparse reward environments in reinforcement learning (RL) pose significant challenges for exploration, often leading to inefficient or incomplete learning processes. To tackle this issue, this work proposes a \textbf{teacher-student RL} framework that leverages Large Language Models (LLMs) as "teachers" to guide the agent's learning process by decomposing complex tasks into subgoals. Due to their inherent capability to understand RL environments based on a textual description of structure and purpose, \textbf{LLMs} can \textbf{provide subgoals} to accomplish the task defined for the environment in a similar fashion to how a human would do. In doing so, three types of subgoals are proposed: positional targets relative to the agent, object representations, and language-based instructions generated directly by the LLM. More importantly, we show that it is possible to \textbf{query the LLM only during the training phase}, enabling agents to operate within the environment without any LLM intervention. We assess the performance of this proposed framework by evaluating three state-of-the-art open-source LLMs (Llama, DeepSeek, Qwen) eliciting subgoals across various procedurally generated environment of the MiniGrid benchmark. Experimental results demonstrate that this curriculum-based approach accelerates learning and enhances exploration in complex tasks, achieving up to \textbf{30 to 200 times faster convergence} in training steps compared to recent baselines designed for sparse reward environments.
\end{abstract}

\section{Introduction}
Let us imagine a scenario where we must master a new skill where feedback is only provided after long periods of effort, i.e., no guidance, no information about the progress, just endless trial and error. This is the experience of many Reinforcement Learning (RL) agents operating in sparse reward environments, where the scarcity of feedback makes the learning process slow and inefficient. In such environments, the traditional approach of random exploration—where an agent learns by trying out different actions and receiving rewards only occasionally—often falls short. The randomness inherent to this exploration strategy poses significant learning challenges, as agents must rely on serendipity to encounter rewarding states. Without frequent or consistent feedback, the agent may spend vast amounts of time exploring irrelevant or ineffective actions, making the learning process highly resource-intensive. Yet, even with impressive achievements \citep{silver2016mastering, mnih2015human}, RL agents still require assistance in environments where rewards are infrequent and learning becomes more challenging.

Just as humans benefit from mentors, RL agents can similarly gain from mentor-like support to explore more efficiently. In traditional approaches, Curriculum Learning (CL) has been employed to guide agents through progressively complex tasks, such as those in MiniGrid \citep{minigrid}, improving learning efficiency during the exploration process. While CL provides a structured learning path \citep{andreas2017modular, sukhbaatar2017intrinsic}, approaches falling in this family of RL algorithms often lack adaptability to new environments and do not inherently grasp the contextual nuances of different tasks.

\begin{figure}
    \centering
    \includegraphics[width=0.7\linewidth]{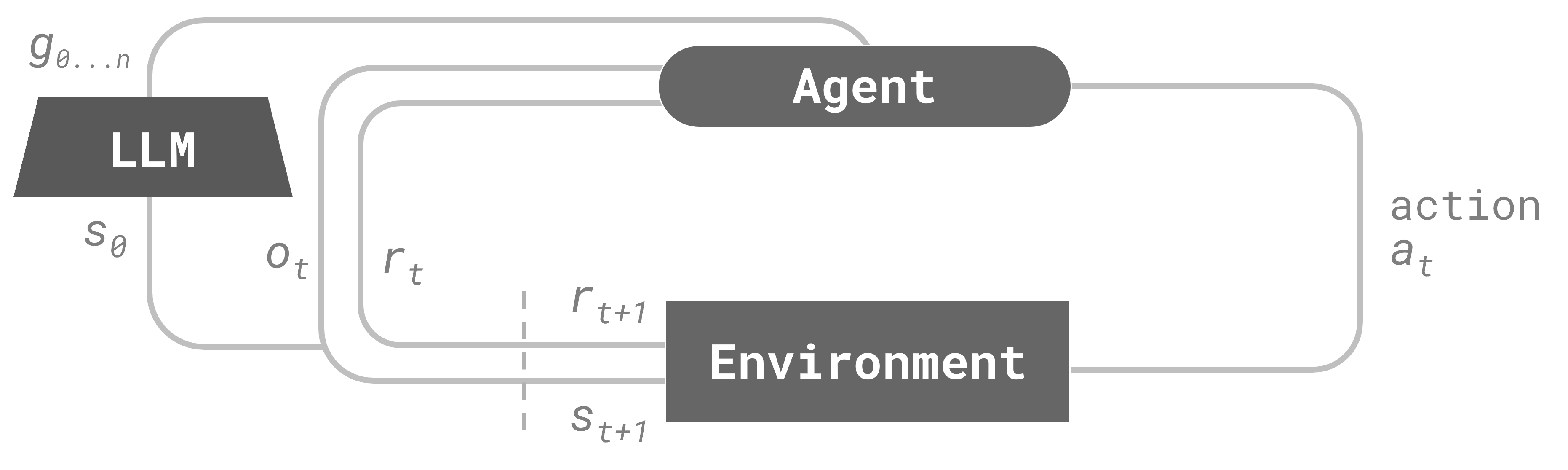}
    \caption{RL framework with teacher LLM. In this framework, $g_0,...,g_N$ are the subgoals provided by the LLM at the beginning of each episode according to the initial state information $s_0$.}
    \label{fig:rl_framework_with_llm}
\end{figure}

In recent years, Large Language Models (LLMs) have shown promise in enhancing RL by providing contextual understanding and adaptive guidance, meaning LLMs can interpret task requirements and adjust suggestions based on the current environment. Leveraging this potential, we propose using LLM as \emph{information processors} \citep{cao2024survey} within a RL framework, adopting a goal-oriented policy $\pi(a_t | o_t, g_n)$ tailored for sparse environments (Figure \ref{fig:rl_framework_with_llm}). Inspired by CL, we adapt its principles to focus on goal decomposition rather than a progression of difficulty. LLMs generate subgoals that break down complex tasks into manageable steps, offering structured feedback and addressing the challenge of sparse rewards.

We evaluate our results in procedurally generated environments from the MiniGrid benchmark \citep{minigrid}, where each environment is designed to assess generalization capabilities by ensuring that each episode differs from the previous one due to random modifications of the environment's components\footnote{This is why the terms "episode" and "level" are used interchangeably.}. We compare the results against recent baselines in teacher-oriented RL training for MiniGrid--AMIGo \citep{campero2020learning} and L-AMIGo \citep{mu2022improving}---achieving up to a 200-fold reduction in the number of training steps required for the agent's effective learning of the RL task.

In summary, our contributions can be summarized as follows: \textbf{(i)} we propose using LLMs as expert teachers to generate subgoals that guide RL agents, enhancing learning efficiency in sparse reward environments; \textbf{(ii)} we introduce a goal-oriented policy framework that leverages these LLM-generated subgoals, significantly outperforming recent baselines in teacher-oriented RL within the MiniGrid benchmark; \textbf{(iii)} we conduct a comprehensive evaluation of multiple state-of-the-art LLMs, identifying which models are most effective for this role; and \textbf{(iv)} we develop a method to model LLMs that reduces prompting times and computational costs, making the training process scale up better to cope with real-world applications.

\section{Related work}
\paragraph{Curriculum Learning (CL)} Organizing learning tasks in a structured sequence has emerged as an effective strategy in training models. Originally conceived as "train from easy to hard", CL has attracted attention for its potential to expedite learning processes \citep{elman1993, bengio2009}. However, its efficacy is not universally consistent across implementations, where the curriculum must be well-defined in order to be effective, as indicated by several studies \citep{graves2017automated, hacohen2019, platanios2019}. These findings suggest that while CL holds promise, its application requires careful considerations.

Curriculum Learning in RL is brought to life through RL teachers, which constitute a dynamic methodology for guiding RL processes. One notable advancement in this area is the work by \citep{florensa2018automatic} who developed a generative adversarial network to create goals of suitable difficulty for the agent. Similarly, \citep{matiisen2019} introduced a teacher that automatically selects subtasks from a predefined set, based on the student's performance, aiming to mitigate the risk of forgetting. \citep{graves2017automated} demonstrate that employing a stochastic syllabus can result in notable improvements in the efficiency of CL. In a purely teacher-student scenario, \citep{racaniere2019automated} introduced a goal-setter network, focusing on feasible, valid and diverse subgoal generation. In contrast, AMIGo \citep{campero2020learning}, introduces an adversarial teacher-student framework. This teacher proposes goals to train a goal-conditioned student policy in the absence of alongside environment rewards. L-AMIGo \citep{mu2022improving}, an extension of AMIGo, introduces language abstractions as subgoals, improving the previous results.

\paragraph{LLM-enhanced RL} Leveraging the power of LLMs has become an emerging strategy in the field of RL, exploiting the capabilities of natural language processing for RL frameworks. Models like the closed-source GPT series \citep{radfordLanguageModelsAre, brown2020language, achiam2023gpt} or open-source alternatives such as Llama \citep{touvron2023llama}, Qwen \citep{bai2023qwen}, and DeepSeek \citep{bi2024deepseek} are nowadays being harnessed to guide RL agents.

Recent surveys in the field \citep{cao2024survey, hu2024survey} have provided substantial insights into the diverse capabilities of language models, highlighting various studies that integrate LLMs into the RL framework. One such approach, where LLMs are employed as decision-makers, is LID \citep{li2022pre}, which transforms goals, histories, and observations into a sequence of embeddings, allowing an LLM to predict the next action. Another remarkable work in this direction is GLAM \citep{carta2023grounding}, which focuses on aligning LLMs with the environment by fine-tuning them to act as direct agents.

LLMs have shown remarkable versatility not just as decision-makers, but also as designers of reward systems. For example, Eureka \citep{ma_eureka_2023} demonstrates the ability of LLMs to iteratively improve reward functions through coding, self-reflection, and RL simulations. In a similar vein, \citep{kwon2023reward} demonstrates how LLMs can generate reward signals from natural language prompts, converting user-provided task descriptions into rewards. Motif \citep{klissarov_motif_2023} hinges on LLMs to create intrinsic rewards from user-defined preferences, resulting in superior performance in the NetHack benchmark. 

Beyond textual data comprehension and generation, LLMs also excel at simulating and navigating complex environments. IRIS \citep{micheli2022transformers} uses a world model combining a picture decoder and a transformer to imagine millions of possible paths, outperforming humans in 10 out of 26 Atari 100k games.

Finally, LLMs can serve as \emph{information processors}, offering new ways to simplify tasks. DEPS \citep{wang_describe_2023} employs GPT-3 for interactive zero-shot planning in Minecraft, where it decomposes tasks into subtasks and adjusts plans based on execution outcomes and error summaries. SayCan \citep{ahn_as_2022} uses an LLM to assign relevance scores to robot actions based on their natural language descriptions and value functions, determining the best action by combining these scores. Jarvis-1 \citep{wang_jarvis-1_2023} employs multimodal language models and memory to generate and execute plans in Minecraft, completing over 200 tasks and surpassing current state-of-the-art agents in Minecraft. ELLM \citep{du2023guiding} leverages LLMs to suggest goals for exploration, guiding agents toward human-meaningful behaviors without needing human input.

\paragraph{Contribution} While AMIGo \citep{campero2020learning}, L-AMIGo \citep{mu2022improving} and our study rely on teacher-student frameworks validated in the MiniGrid benchmark, AMIGo and L-AMIGo require its teacher to generate subgoals from scratch due to its adversarial setup. Differently, our approach leverages a pretrained teacher—specifically, a LLM—to produce these subgoals.
Akin to ELLM where subgoals are proposed with LLMs, ELLM focuses on maximizing the diversity of skills acquired by the agent, with subgoals that are not necessarily sequential or task-specific. In contrast, our approach focuses on time-sensitive, curriculum-based subgoals that guide the agent through a structured sequence, enhancing learning efficiency and effectiveness. Given these fundamental differences, comparing our approach directly against ELLM is not meaningful, as the underlying methodologies and objectives are distinct.

\section{Methodology}
As depicted in Figure \ref{fig:rl_framework_with_llm}, our proposed goal-oriented policy framework introduces a subgoal $g_n$, provided by the LLM, to the observation $o_t$ and the reward signal $r_t$. Acting as an omniscient supervisor, the LLM has full access to the environment's initial state $s_0$. At the start of each episode, it generates a sequence of subgoals $g_0, g_1,\ldots,g_N$, where $N+1$ represents the total number of subgoals. During the episode, in each time step, the agent receives its current observation $o_t$, the reward signal $r_t$, and the relevant subgoal $g_n$ from this sequence. Once a subgoal $g_n$ is achieved, the agents transitions to the next subgoal in the sequence $g_{n+1}$ at the subsequent time step $t+1$. This transition implies that $g_{n+1} \neq g_n$, ensuring that the agent progresses through a sequence of unique subgoals.

In this framework, the RL agent learns a policy $\pi(a_t|o_t,g_n)$ designed to maximize the expected cumulative reward \smash{$\sum_{k=0}^{T}\gamma^k\cdot R_{t+k+1}(s_t|g_n)$} given both the observation and the current subgoal.

\subsection{Subgoal generation}
Subgoal generation takes place at the beginning of each episode, where the LLM, with access to the complete environment state $s_0$, creates a sequence of subgoals $\{g_0,g_1,g_2,\ldots,g_N\}$, which progressively guides the agent towards the final goal of the RL task, reducing the reward horizon.
In addition to the extrinsic reward $r_t^e$ coming from the tasks at hand, we introduce an auxiliary reward $r_t^g$ that incentives the achievement of subgoals. Therefore, the reward that the agent receives in each time step is given by: 
$r_t = r_t^e + \alpha \cdot r_t^g$, where $\alpha\in\mathbb{R}^+$ is a scaling hyperparameter that balances the weight of subgoal completion with respect to the overall task achievement. The reward $r_t^g$ for each subgoal $g\in\{g_1,\ldots,g_N\}$ is defined as:
\begin{equation}
    r_t^g = 1 - \frac{\text{t}_g}{\text{t}_{max}},
\end{equation}
where $\text{t}_g$ denotes the number of steps taken to achieve a specific subgoal, and $t_{max}$ represents the maximum number of steps allowed in each episode, which acts as a normalization component.


We propose providing subgoals to the agent in three ways: as positions, as representations, and as language embeddings.

\textbf{Position-based Subgoals:} These involve defining $\{x,y\}$ coordinates to indicate locations of the predicted subgoals, focusing on relative positions due to the agent's lack of absolute awareness. This method was expected to be effective, but the agent's dependence on relative directions can make long-term navigation challenging. Additionally, LLMs, which are not optimized for this task, can struggle with accurately identifying object locations within 2D grids, leading to potential misalignments.

\textbf{Representation-based Subgoals:} These subgoals use a grid encoding scheme to reflect how objects are perceived by the agent. This approach aligns subgoals with the agent’s internal representation of the environment, allowing for unique and semantically meaningful identification of objects. While this method can enhance interaction with specific features, reliance on a fixed grid scheme might limit the agent's adaptability to new, potentially diverse environments. For example, in an open-world scenario, a new type of object might appear.

\textbf{Language-based Subgoals:} These subgoals utilize embeddings generated from a Language Model, converting LLMs text output into embeddings to provide guidance in a more abstract and flexible form. This approach offers the potential for high adaptability and generalization across diverse tasks and environments. However, due to the inherently stochastic nature of LLMs, the output may vary from one instance to another. This variability can result in different phrasings for the same subgoal, effectively expanding the pool of potential subgoals.

\subsection{Prompt engineering}
Prompt engineering is crucial for optimizing the effectiveness of LLMs \citep{brown2020language, reynolds2021prompt}, with Chain of Thought (CoT) being a mainstream technique for enhancing the performance of LLMs. CoT guides LLMs to articulate their reasoning process, thereby improving their ability to generate relevant subgoals by breaking down complex tasks into logical steps. In our study, we employ Zero-shot CoT \citep{kojima2022large} to ensure that subgoal generation is driven by the LLM’s inherent reasoning rather than biased by example-based prompts. This approach allows extracting unbiased subgoals, which in addition will turn to be useful when effectively modeling the output of the LLM at inference time. We refer the reader to \hyperlink{apA}{Appendix A} for detailed information about the specific prompts used in our work.

\subsection{Statistical modeling of LLMs} \label{sec:modeling}

Our initial approach uses LLMs to generate subgoals at the beginning of each episode to guide the agent through complex tasks. However, training in PCG environments often requires thousands to millions of episodes/levels. Relying on the LLM to generate subgoals every time a new episode begins may not be scalable when dealing with such a large number of episodes, due to resource-intensive factors like latency, hardware requirements, and API costs (e.g., when using services like ChatGPT). To mitigate these scalability issues,  \textbf{we propose an offline subgoal modeling method}, shown in Algorithm \ref{alg:subgoals}.


In our proposed method, prior to the training stage, we prompt the LLM  with the initial state $s_0$ across a randomly selected subset of levels from the environment's entire level distribution. This process results in $M$ levels, from which we collect a set of subgoal proposals for each level.
To provide optimal guidance for the agent, we assume access to an Oracle (e.g., a human expert) who can decompose each level and propose optimal subgoals. These Oracle-provided subgoals are meaningful and logical within the environment, aligning perfectly with the overall task objectives.
By comparing these optimal subgoals to the LLM's proposals, we model the discrepancies (i.e., errors) between the LLM's output and the desired subgoal decomposition. 
This allows us to build a statistical model that generates subgoals for any level in the environment's distribution during training, enabling the agent to train over the entire level distribution—not just the $M$ levels used for subgoal collection. As a result, we eliminate the need to query the LLM over hundreds of thousands of episodes during the training phase.


\begin{tcolorbox}[title=Algorithm 1: Subgoal Evaluation] 
\begin{algorithm}[H]
\label{alg:subgoals}
$\mathcal{G}_{n}^m$: The set of subgoal proposals at iteration $m$. \\
$\mathcal{O}_{n}^m$: The set of optimal subgoals at iteration $m$. \\
\For{$m = 1$ to $M$}{
    Prompt the LLM for subgoal proposals $\mathcal{G}_{n}^m$ in level $L^m$. \\
    Collect set of subgoal proposals $\mathcal{G}_{n}^m = \{g_{0}^m, g_{1}^m, \ldots, g_{N}^m$\}. \\
    Label the optimal subgoal proposals $\mathcal{O}_{n}^m$ for each level $L^m$ using the Oracle. \\
}
Model the error between the optimal subgoal $\mathcal{O}_{n}^m$ and the LLM proposals $\mathcal{G}_{n}^m$.
\end{algorithm}
\end{tcolorbox}

\begin{figure}[t]
    \centering
    \includegraphics[width=0.5\linewidth]{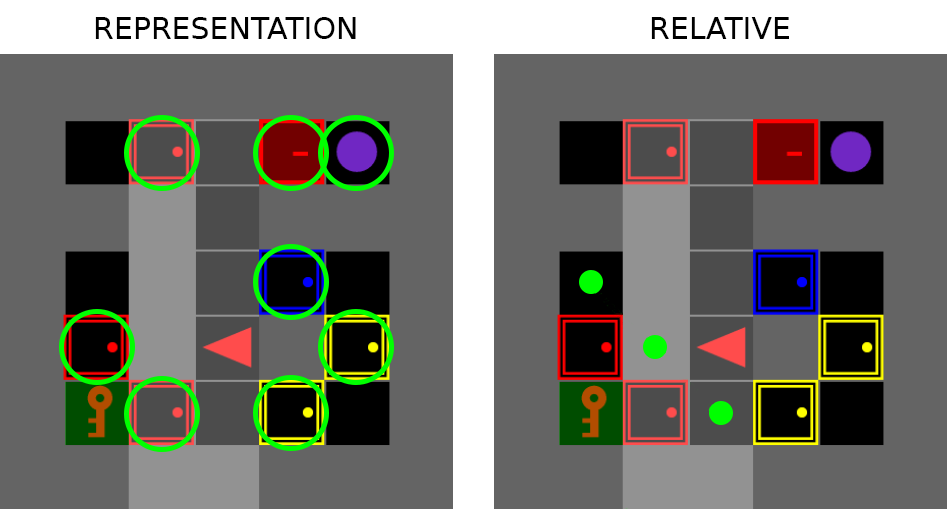}
    \caption{An episode belonging to the environment KeyCorridorS3R3 from MiniGrid, where the possible subgoals subject to the error are shown.The first subgoal is the key located at the bottom left position of the grid. (left) Possible subgoals for representation-based subgoals; (right) the possible grid locations according to a Manhattan distance of 2.}
    \label{fig:manhattan_error}
\end{figure}

To evaluate the accuracy of subgoal generation, we employ different metrics tailored to each type of subgoal. For \emph{representation-based} subgoals, we measure accuracy by how often the LLM's proposed subgoals match the target subgoals in terms of object type, color, state, and sequence. Using this calculated accuracy, we model errors during training by intentionally selecting alternative objects within the environment that are not the intended subgoals, thereby reflecting the types of mistakes LLMs sometimes make (see Figure \ref{fig:manhattan_error}-left).

For \emph{relative-based} subgoals, which focus on positional accuracy, we use metrics such as overall accuracy, the mean Manhattan distance error (the average distance between the proposed and optimal subgoals), and the standard deviation of the Manhattan distance error (reflecting the variability in distance errors). Based on these measurements, we model positional inaccuracies during training by choosing different points within the grid—including infeasible subgoals like walls—to mirror the LLM's performance discrepancies, as illustrated in Figure \ref{fig:manhattan_error}-right.

Lastly, due to the complexity involved in modeling \emph{language-based} subgoals, we limit the training process to the same set of $M$ levels used during the offline statistical modeling rather than the actual level distribution. In this case, we do not model the error, but directly utilize the LLM's outputs for each level.


Despite the need for an Oracle to model these discrepancies, relying exclusively on the Oracle for subgoal generation during training is impractical in real-world scenarios due to resource constraints and scalability issues. Training the agent solely on Oracle-provided subgoals could offer perfect guidance, but at the risk of a penalty in performance if inaccuracies arise when LLM's are used during deployment. By leveraging the LLM --despite its susceptibility to errors-- and introducing stochasticity through our statistical model, we enable the agent to learn from variability inherent in LLM outputs. This approach not only reduces dependence on the Oracle but also ensures the agent develops robustness and adaptability necessary for real-world applications.

\section{Experiments and results}

To evaluate the effectiveness of our proposed method, we conduct experiments across several challenging environments within the MiniGrid framework \citep{minigrid}: \textbf{MultiRoomN2S4}, \textbf{MultiRoomN4S5}, \textbf{KeyCorridorS3R3}, \textbf{KeyCorridorS6R3}, \textbf{ObstructedMaze1Dlh} and \textbf{ObstructedMaze1Dlhb}. We use an implementation 
\citep{Andres_2022} of Proximal Policy Optimization (PPO) \citep{schulman2017proximal} to test our LLM-generated subgoals. For detailed information on the selected hyperparameters and model architecture, please refer to \hyperlink{apC}{Appendix C}.

In our experimentation, we use the modeled LLMs introduced in Section \ref{sec:modeling} to generate subgoals at the beginning of each episode. The LLM receives a full observation of the environment, $s_0$, to generate the subgoals, while, the agent has access only to a $7\times7\times3$ observation tensor representing the area directly ahead. We evaluate the performance of three open-source LLM models: Llama3-70b \citep{touvron2023llama}, Qwen1.5-72b-chat \citep{bai2023qwen}, and DeepSeek-coder-33b \citep{bi2024deepseek}, and analyze their impact during training when using each type of proposed subgoals---representation, relative, language.

To improve scalability and practicality, we further investigate the effects of omitting rewards and subgoals during training using the top-performing LLM model and the most effective type of subgoal identified in previous analyses. We evaluate two scenarios: (1) \textbf{no reward}, where the agent receives subgoals but no additional rewards for achieving them, preserving the reward horizon for assessing performance without immediate rewards; and (2) \textbf{no subgoal}, where subgoals are excluded from the agent's observation space, but rewards are still provided upon subgoal completion.

\begin{figure}[t]
    \centering
    \includegraphics[width=\linewidth]{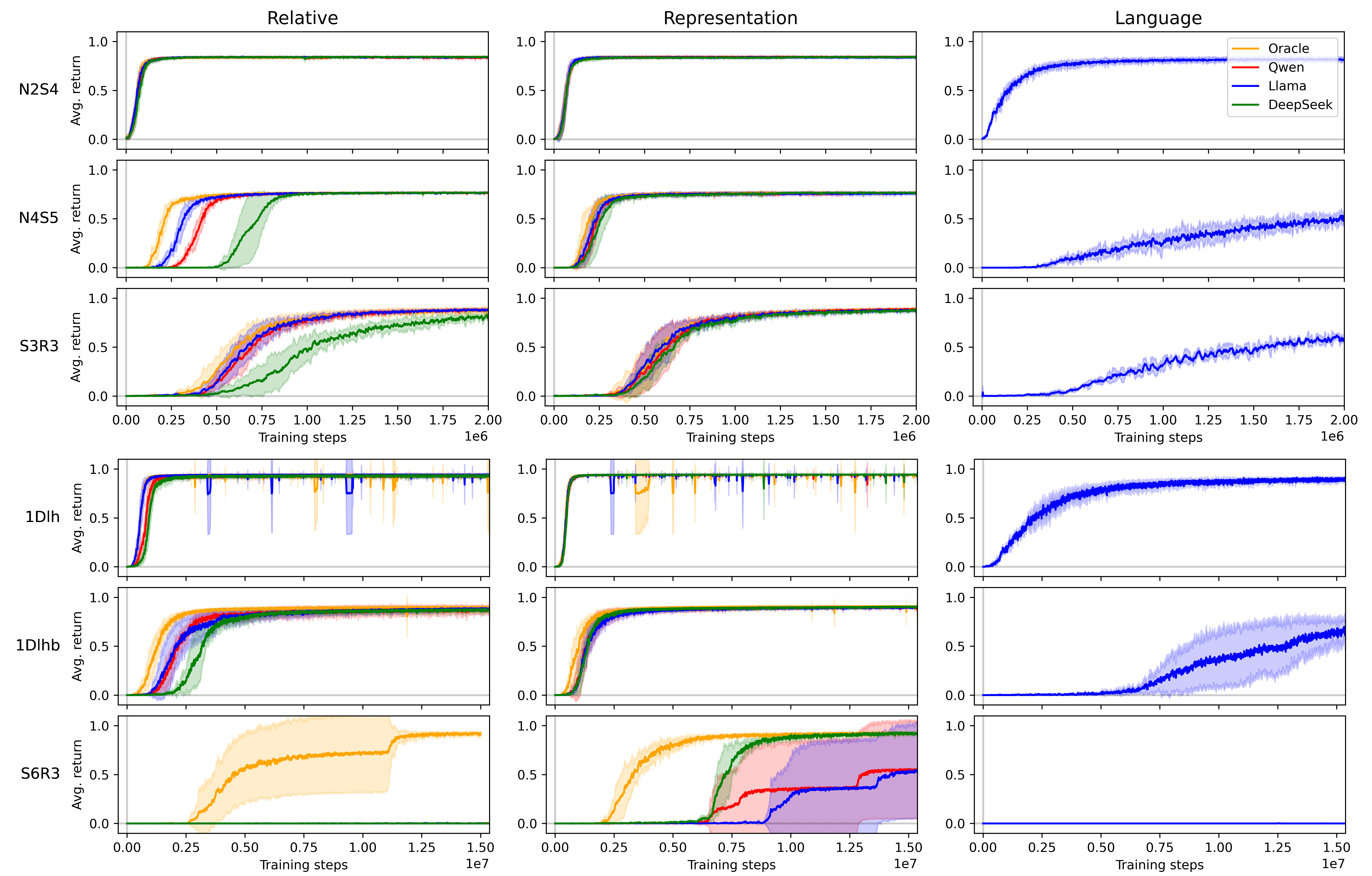}
    \caption{Comparison of the training curves for the proposed CL-based LLM-assisted RL training framework, showing the average return over training steps for the six environments using three methodologies: relative to the agent's position (first column), representation-based (second column) and language-based (third column). The analysis includes three LLMs—Qwen, Llama, and DeepSeek—against the Oracle subgoals. The shaded area represents the variability in the average return across $5$ runs of the agent's training process.}
    \label{fig:rq3}
\end{figure}

\paragraph{Comparison between LLMs} As shown in Figure \ref{fig:rq3}, Llama consistently outperforms the rest of models across the majority of environments. In contrast, DeepSeek exhibits the lowest performance across many of the evaluated environments. Despite the significant differences in error modeling across LLMs, the performance disparity is less noticeable in relatively simpler environments. In fact, the choice of subgoal type appears to have a more substantial impact on performance than the specific LLM model in use. For detailed results on the LLM modeling (Section \ref{sec:modeling}), we refer to \hyperlink{apD}{Appendix D}

\paragraph{Comparison between subgoal representation strategies} Representation-based subgoals consistently outperform relative-based subgoals, particularly in more complex environments. For instance, in KeyCorridorS6R3, even with the Oracle’s perfect subgoals, the agent struggles to learn when following relative subgoals. This is because the agent often encounters obstacles, like walls, when trying to follow directional instructions, forcing it to explore alternative routes. In contrast, representation-based subgoals align more effectively with the agent’s internal perception of the environment, leading to more accurate navigation and faster task completion.

Surprisingly, even if DeepSeek obtains the lowest accuracy when modeling its error in KeyCorridorS6R3 representation-based subgoals, it outperforms the other models in terms of training effectiveness. Although counterintuitive, we hypothesize that this is due to the longer reward horizon in this task. Since the subgoals are spread across a larger grid, DeepSeek’s errors sometimes inadvertently place the subgoal closer than the actual subgoal. When the agent follows these incorrect subgoals, it still receives small rewards, reinforcing its tendency to follow the LLM's guidance. This positive reinforcement accelerates the learning process, despite the initial inaccuracy.

We end our discussion related to subgoal representation strategies by focusing on the case of language-based subgoals produced by Llama, which outperformed other LLM models\footnote{This choice was made due to computational demands.}. Embedding Llama's output through a Language Model, specifically all-MiniLM-L6-v2, introduces challenges. The larger embedding size and the broader subgoal space make it more difficult for the agent to process the elicited subgoal descriptions and learn effectively.

\paragraph{Impact of Reward Balance} After evaluating the impact of reward, we opted for a low $\alpha$ value and a normalized reward by the steps taken to achieve each subgoal. A high value of $\alpha$ places too much weight on subgoal completion, causing the agent to overly rely on subgoal accuracy, which is not always reliable given the current results of our prompting strategy. By using a lower $\alpha$, the agent is guided by subgoals but still prioritizes the overall RL task, ensuring that subgoals serve as helpful aids rather than the main focus. Furthermore, normalizing the reward by the steps taken to achieve each subgoal encourages quicker subgoal completion.

\begin{figure}
    \centering
    \includegraphics[width=\linewidth]{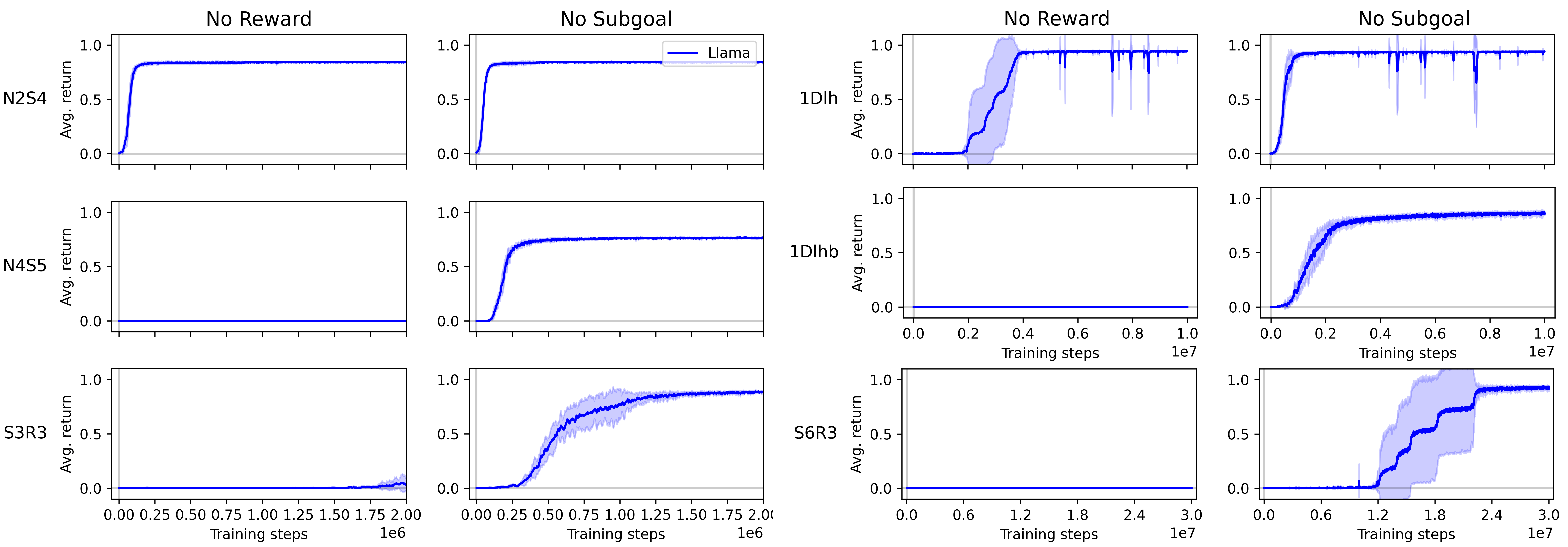}
    \caption{Performance comparison for different training scenarios using Llama and representation-based subgoals. The figure displays results for no-reward training (first and third columns) and no-subgoal training (second and fourth columns). The no-reward training condition shows the agent's performance when rewards are absent, but representation subgoals are present, while no-subgoal training condition shows the performance when subgoals are absent, but rewards are still present.}
    \label{fig:rq4}
\end{figure}

\paragraph{Ablation study on the impact of subgoals and rewards on agent training} Building on previous findings, using Llama as the LLM model and representation-based subgoals, we explore further by analyzing two specific training conditions. In the "no reward" condition, where only representation-based subgoals are used and supplementary rewards for achieving rewards these subgoals are omitted (depicted in the first and third column of Figure \ref{fig:rq4}) the agent fails to learn how to solve the problem. This highlights the critical role of rewards in effective training, especially scenarios with a high reward horizon. In contrast, in environments with a shorter reward horizon, such as MultiRoomN2S4 and ObstructedMaze1Dlh, the agent successfully learns the task under this setup. Conversely, in the "no subgoal" condition, where supplementary rewards are provided without including subgoals in the observations (depicted in the second and fourth column of Figure \ref{fig:rq4}) the agent performs well and converges similarly to when subgoals are provided.

These results indicate that while LLMs can be valuable for guiding the agent during training, accelerating considerably their training times, their presence is not necessary once training is complete, reducing computational overhead and operational costs of the agent in deployment.

\section{Conclusions}
This work has evinced that LLMs can be harnessed as subgoal generators, to yield a valuable strategy for training RL agents to tackle tasks characterized by reward sparsity. By applying curricular learning techniques, we can significantly \textbf{reduce the reward horizon}, accelerating the training process. The knowledge embedded in \textbf{LLMs facilitates faster and more effective training}. Furthermore, if trained without including the subgoals in the observation, \textbf{the agent does not need to query the LLM during the deployment phase}, reducing computational overhead and costs.

Our proposed framework, while effective, has some limitations and avenues for future research. Currently, when it comes to language-based subgoals, we rely on direct text output by the LLM, which can vary significantly and may present challenges in managing the volume of possible subgoals. Future efforts will focus on filtering these outputs against a predefined pool of subgoals, aiming to reduce ambiguities and redundancies observed in the linguistic description of subgoals elicited by the LLM and ultimately, improve the agent's guidance through the produced subgoals. Additionally, extending our approach to a broader range of benchmarks will help evaluate its generalizability. Another key area for future development is designing a more adaptive teaching strategy that can dynamically adjust to the agent's evolving capabilities. For instance, as observed when using DeepSeek, there are still limitations due to insufficient reduction in the reward horizon in some environments. Lastly, investigating the effects of strict adherence to LLM guidance could further enhance the agent's generalization across diverse tasks, when encountering new tasks that are similar to those previously learned by the agent.

\bibliographystyle{plainnat}
\bibliography{mybibfile}

\newpage
\appendix

\section{Prompts used for subgoal generation}
\hypertarget{apA}
In this appendix, we detail the prompts used to instruct the LLM for subgoal generation at the start of each episode. These prompts are provided sequentially, using a CoT approach to maintain context and build upon previous responses.

Figure \ref{fig:prompt_1} illustrates the initial prompt, where the LLM is asked about its knowledge of MiniGrid. This step establishes a foundational understanding of the environment, which is crucial for generating relevant subgoals.

\begin{figure}[h]
\begin{tcolorbox}[title=Subgoal Generation Prompting] 

\centering
    \raggedright
        \textbf{User:} \\
    
        Tell me about your knowledge of MiniGrid, Farama-Foundation's grid world environments for reinforcement learning. \\              
\end{tcolorbox}
\caption{Initial prompt to the LLM about its knowledge of MiniGrid, establishing a foundational understanding of the environment.}
\label{fig:prompt_1}
\end{figure}

Similarly, Figure \ref{fig:prompt_2} shows a prompt that provides detailed information about the MiniGrid environment, including the encoding of different objects and the representation of the agent’s direction. This prompt ensures that the LLM has a clear understanding of the environment's state and object semantics.

\begin{figure}[h]
    \begin{tcolorbox}[title=Subgoal Generation Prompting]
    \centering
    \raggedright
        \textbf{User:} \\
    
        Each environment in MiniGrid is defined by a grid. I will provide you a matrix representing this grid. \\ \vspace{7pt}
        
        Each tile in the grid is encoded with a number, and each number represent the following: \\ \vspace{7pt}
        
        0: Wall, walls are not movable objects, that cannot be overpassed. \\
        2: Key, the key is used to unlock locked doors. \\
        x: Floor, the agent can walk of these. \\
        3: Ball, the ball is the final goal of the environment. \\
        7: Closed door, these doors can be opened without a key. \\
        8: Locked door, these doors need a key to be opened. \\
        9: Opened door, these doors can be overpassed. \\ \vspace{7pt}
        
        Also the agent is represented by: \\
        >: Looking right \\
        V: Looking down \\
        <: Looking left \\
        \textasciicircum: Looking up \\ \vspace{7pt}
        
        With all of this in mind, describe the semantics of the state, using only information from the state and your knowledge of MiniGrid. \\       
    \end{tcolorbox}
    \caption{Prompt providing detailed information about MiniGrid's grid encoding scheme and object representation, ensuring the LLM understands the state semantics.}
    \label{fig:prompt_2}
\end{figure}

Figure \ref{fig:prompt_3} presents the final prompt, where the LLM is explicitly instructed to generate subgoals based on the current state. The LLM is guided to focus on specific objects of interest rather than movement patterns and is asked to provide the subgoals in a structured format.

\begin{figure}[h]
    \begin{tcolorbox}[title=Subgoal Generation Prompting]
    \centering
    \raggedright
        \textbf{User: } \\
    
        Now, respond by explicity declaring which subgoals would be optimal to progress towards the goal. Remember that keys are needed to open locked doors. \vspace{7pt}

        Do not provide subgoals related to movement patterns. Instead, focus on subgoals involving specific objects of interest. Also, provide the final goal as subgoal. \vspace{7pt}
        
        For the subgoals, follow the following format: \\
        <subgoals> \\
        1. instruction (str) (x (int), y (int)) \\
        2. instruction (str) (x (int), y (int)) \\
        3. instruction (str) (x (int), y (int)) \\
        </subgoals> \\
        The instruction must be text without any number. The x and y are the coordinates of the subgoal. Do not add any more information. \vspace{7pt}
        
        The state the agent is seeing is the following: \{state\}  
    \end{tcolorbox}
    \caption{Final prompt instructing the LLM to generate specific subgoals based on the current state, focusing on object-related goals and providing subgoals in a structured format.}
    \label{fig:prompt_3}
\end{figure}

These prompts are delivered sequentially at the first time step of each episode in state $s_0$, with the LLM’s responses building on the information provided in the previous prompts. This method leverages Chain of Thought reasoning, allowing the LLM to maintain memory and coherence throughout the prompting process.

\section{Subgoal types with illustrative examples}
\hypertarget{apB}
In this appendix, we present examples of the different types of subgoals used throughout our experiments, including relative, representation-based, and language-based subgoals.

\subsection{Relative-based subgoals}

\begin{figure}[h]
    \centering
    \includegraphics[width=0.5\linewidth]{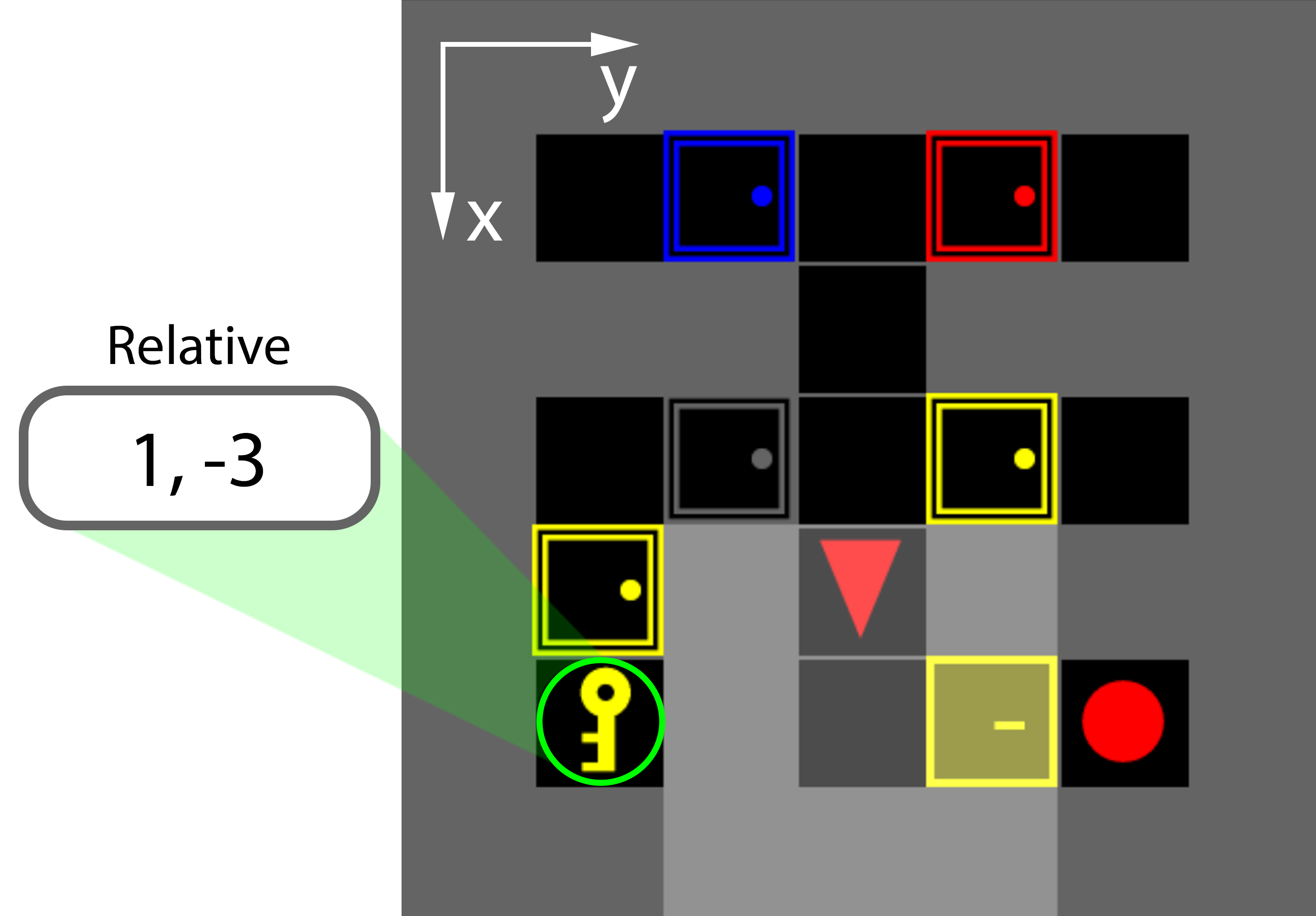}
    \caption{A level from the KeyCorridorS3R3 environment in MiniGrid, where the agent's subgoal is to retrieve a yellow key located in the bottom-left corner of the level. The subgoal is provided as a relative position, with the first value indicating the distance along the x-axis and the second value representing the distance along the y-axis from the agent's current position. }
    \label{fig:relative_ej}
\end{figure}

As shown in Figure \ref{fig:relative_ej}, relative subgoals are represented as a 2-dimensional vector. The first value corresponds to the distance along the x-axis, while the second value represents the distance in grid blocks along the y-axis. With this type of subgoals, the agent learns to interpret the direction it must move in to reach the subgoal, adjusting its navigation based on the relative position provided.

\subsection{Representation-based subgoals}
\begin{figure}[h]
    \centering
    \includegraphics[width=0.5\linewidth]{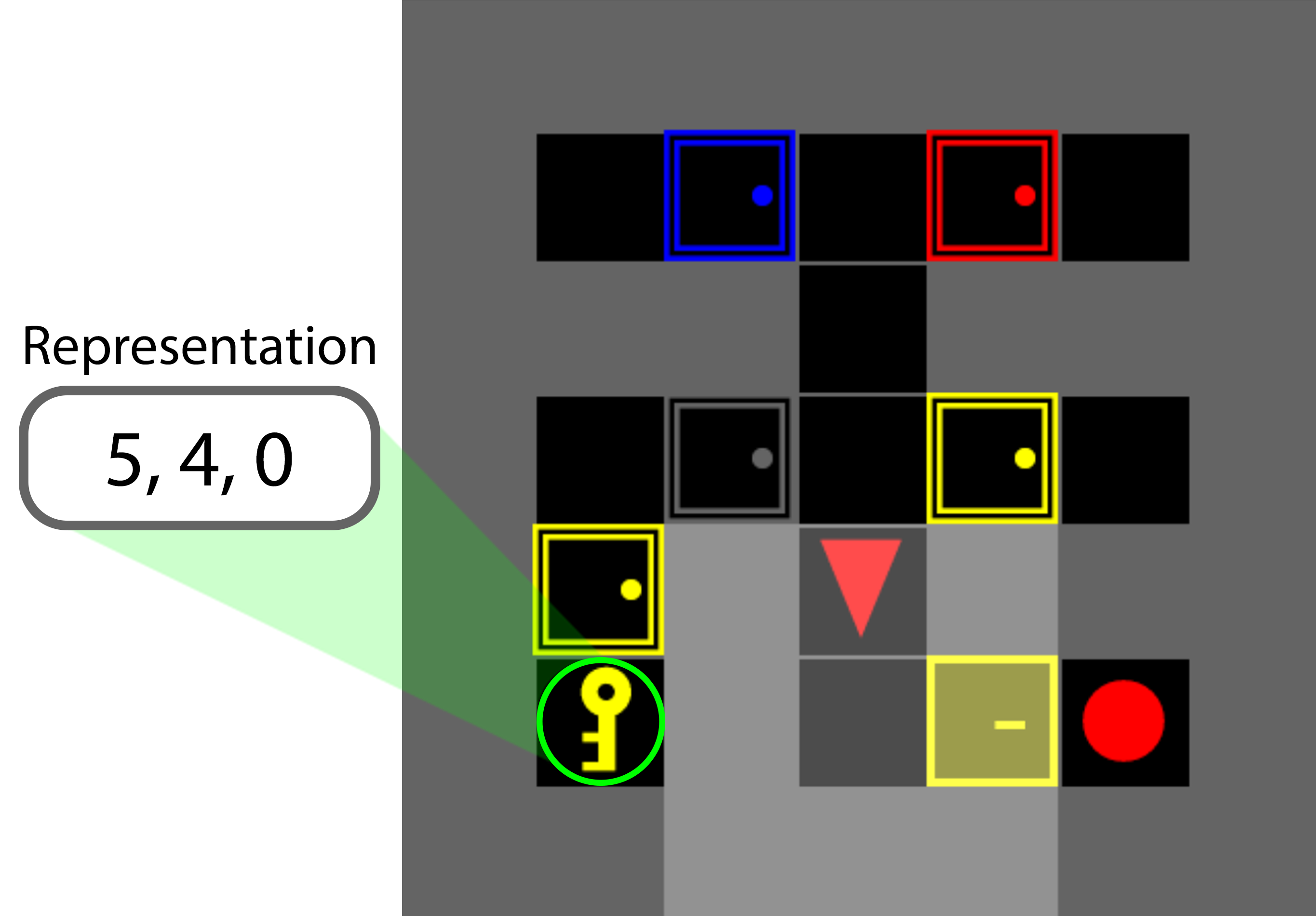}
    \caption{A level from the KeyCorridorS3R3 environment in MiniGrid, where the agent’s subgoal is encoded as a 3-dimensional vector representing a specific object. The first value denotes the type of object, the second value indicates the object's color, and the third value describes its state (relevant for doors).}
    \label{fig:representation_ej}
\end{figure}

As shown in Figure \ref{fig:representation_ej}, representation subgoals are encoded as a 3-dimensional vectors. The first value denotes the type of object, the second represents the object's color, and the third indicates its state. There are 11 possible object types, including unseen (0), empty (1), wall (2), floor (3), door (4), key (5), ball (6), box (7), goal (8), lava (9), and agent (10). Since representation subgoals are limited to interactable objects within the environment, the subgoal space is restricted to doors, keys, balls, boxes, and goals. For colors, six options are available: red (0), green (1), blue (2), purple (3), yellow (4), and grey (5). The state value is particularly relevant for doors, with options for open (0), closed (1), and locked (2).

\subsection{Language-based subgoals}
\begin{figure}[h]
    \centering
    \includegraphics[width=0.5\linewidth]{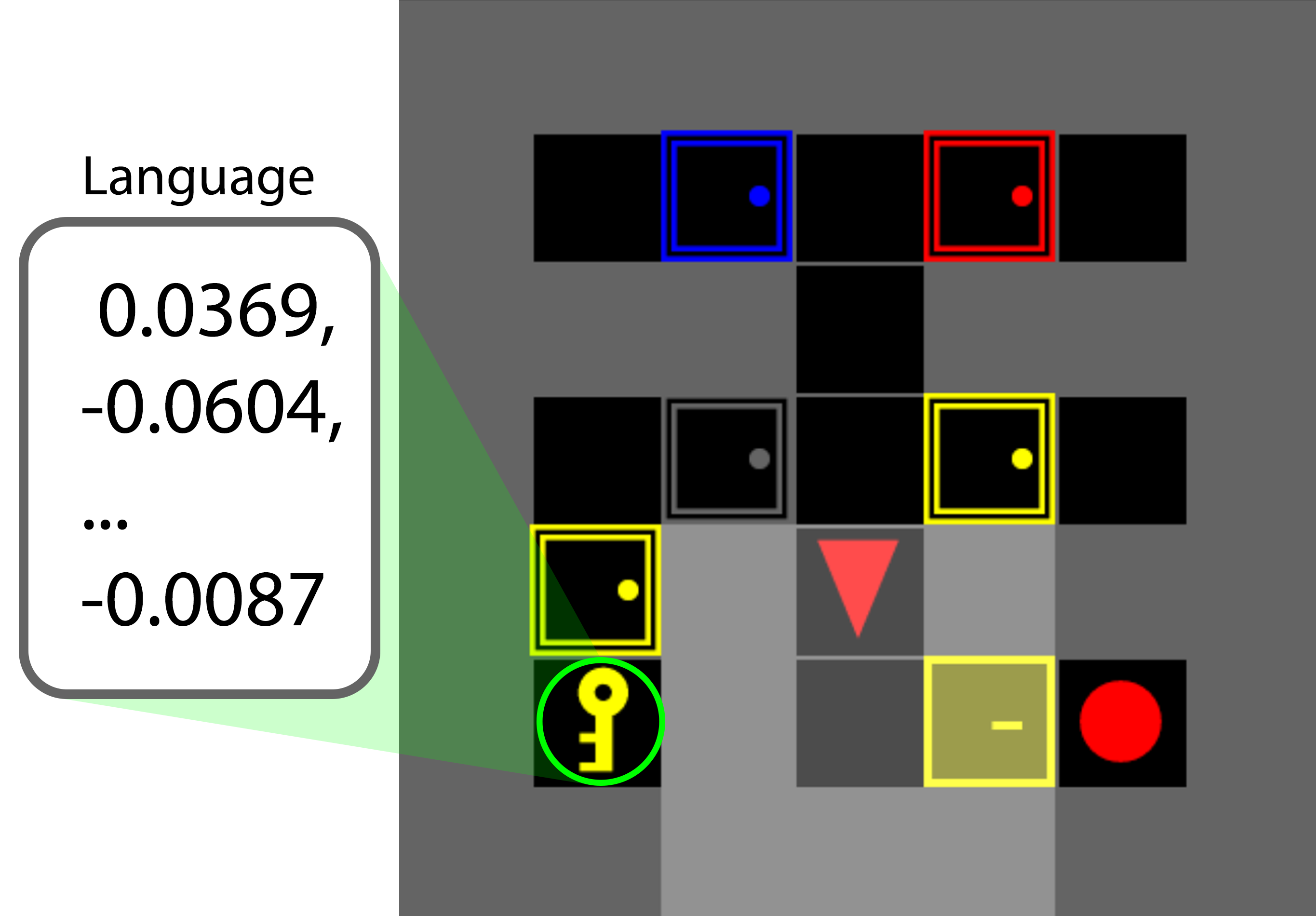}
    \caption{A level from the KeyCorridorS3R3 environment in MiniGrid, where the agent’s subgoal is represented as an embedding generated by a Language Model. The embedding, derived from the output text such as "Go to the key," is a 384-dimensional vector that captures the semantic meaning of the instruction.}
    \label{fig:language_ej}
\end{figure}

As shown in Figure \ref{fig:language_ej}, language-based subgoals are represented as embeddings retrieved from a Language Model based on the output text of the LLM. We use all-MiniLM-L6-v2 as the Language Model, which generates a 384-dimensional vector for each subgoal. For instance, in the subgoal illustrated in Figure \ref{fig:language_ej} the embeddings correspond to the output text "Go to the key." The resulting embedding size is significant, and the subgoal space is extensive since the output is directly taken from the LLM.

\newpage

\section{Model architecture and hyperparameter configuration}
\hypertarget{apC}
This appendix provides detailed information on the architecture and hyperparameters used in our experiments. The algorithm employed for training was Proximal Policy Optimization (PPO), combined with a Convolutional Neural Network (CNN) architecture designed to effectively handle the state representation in MiniGrid environments.

\subsection{Model architecture}
The architecture of our actor-critic model consists of a series of convolutional and fully connected layers. Initially, the model employs three convolutional layers to extract features from the grid observation. Each convolutional layer is defined with 32 filters, a $3\times3$ kernel, a stride of $2\times2$, and padding of $1$, followed by an ELU activation layer.

Following the convolutional layers, the model includes a shared fully connected layer that maps the 32-dimensional output from the convolutional layers to a 256-dimensional vector. This shared representation is then processed by two separate, fully connected networks: one for the actor and one for the critic. The actor network uses a fully connected layer to map the 256-dimensional vector to a vector of size 7, corresponding to the number of possible actions. Meanwhile, the critic network uses another fully connected layer to map the 256-dimensional vector to a scalar value representing the estimated value of the current state.

\subsection{Hyperparameters values}
Table \ref{tab:hyperparameters} summarizes the hyperparameters used in our PPO implementation. Each hyperparameter plays a specific role in shaping the learning process:

\begin{table}[h]
    \begin{center}
    \caption{Hyperparameter values used for PPO in our experiments.}
    \label{tab:hyperparameters}
    \begin{tabular}{clc}
    \toprule
    \textbf{Hyperparameter} & \textbf{Description} & \textbf{Value} \\
    \midrule
    $\alpha$ & Subgoal magnitude & 0.2 \\
    $\gamma$ & Discount factor & 0.99 \\
    $\lambda_{\text{GAE}}$ & Generalized Advantage Estimation (GAE) factor & 0.95 \\
    $\text{PPO rollout length}$ & Number of steps to collect in each rollout per worker & 256 \\
    $\text{PPO epochs}$ & Number of epochs for each PPO update & 4 \\
    $\text{Adam learning rate}$ & Learning rate for the Adam optimizer & 0.0001 \\
    $\text{Adam}\ \epsilon$ & Epsilon parameter for numerical stability & 1e-8 \\
    $\text{PPO max gradient norm}$ & Maximum gradient norm for clipping & 0.5 \\
    $\text{PPO clip value}$ & Clip value for PPO policy updates & 0.2 \\
    $\text{Value loss coefficient}$ & Coefficient for the value function loss & 0.5 \\
    $\text{Entropy coefficient}$ & Coefficient for the entropy term in the loss function & 0.0005 \\
    \bottomrule
    \end{tabular}
    \end{center}
\end{table}

The hyperparameter values in this study were selected based on the findings from \cite{Andres_2022}, where they were meticulously tuned for MiniGrid tasks, ultimately identifying an optimal configuration. By adopting this configuration, we aim to leverage proven settings that have demonstrated success in similar environments, thereby enhancing the reliability and performance of our PPO implementation.

\newpage

\section{Statistical modeling results}
\hypertarget{apD}
We conducted a comprehensive evaluation by querying three LLMs--—Llama3, DeepSeek, and Qwen1.5--—across 1000 distinct instances or levels in each of the six environments. This extensive querying allowed us to gather and analyze their performance metrics, including accuracy and error rates.

Table \ref{tab:rq2} presents a detailed comparison of these LLMs across the six environments. The metrics used for evaluation include:
\begin{itemize}[leftmargin=*]
    \item \textbf{Accuracy:} Measures the percentage of correct predictions made by each LLM out of the total predictions.
    \item \textbf{Correct SGs/Total:} Indicates the number of correct subgoals (SGs) identified by the LLM compared to the total number of subgoals it was asked to predict.
    \item \textbf{Correct EPs/Total:} Measures the number of complete correct episodes (EPs) predicted by the LLM relative to the total number of episodes.
    \item \textbf{Manhattan Dist.:} Represents the average Manhattan distance between the predicted and actual positions, including the standard deviation, providing a measure of spatial accuracy.
\end{itemize}

For position subgoals, a subgoal is considered correct if the specified position matches the actual position of the object. For example, a prediction like "pickup key (2,3)" is correct if a key is indeed located at (2,3). If the predicted position does not match, the error is quantified using the Manhattan distance between the proposed and actual subgoal positions.

In contrast, for representation subgoals, a subgoal is deemed correct if the object name (and color, if relevant) is accurately described. For instance, "pickup key (2,3)" would be correct if the object "key" is mentioned in the description. Errors in representation are not directly quantified in this phase but are addressed in subsequent modeling stages by selecting alternative objects from the grid.

Llama consistently outperforms the other models in both position and representation subgoals. Conversely, DeepSeek, with its simpler 33 billion parameter architecture, often shows the lowest performance compared to Llama's 70 billion parameters.

\begin{table*}[!htb]
	\begin{adjustwidth}{0cm}{0cm}
	\scriptsize
	\setlength{\extrarowheight}{2pt}
    \caption{Performance comparison of LLMs across different environments. The table compares the performance of three language models (Llama, DeepSeek, and Qwen) across six different environments in both Position and Representation modes. The metrics used for comparison include accuracy, the number of correct subgoals (SGs) out of the total SGs, the number of correct episodes (EPs) out of the total EPs, and the average Manhattan distance with standard deviation. The results demonstrate the relative strengths and weaknesses of each model in comprehending and navigating grid-based environments. The model with the highest average performance in each environment is highlighted in bold.}
    \label{tab:rq2}
    \centering
	\resizebox{\columnwidth}{!}{\begin{tabular}{lrrrrr}
		\toprule
		\textbf{Environment} & \textbf{LLM} & \textbf{Accuracy} & \textbf{Correct SGs / Total} & \textbf{Correct EPs / Total} & \textbf{Manhattan Dist.} \\
		\midrule
		 \multirow{3}{*}{\shortstack[c]{\textbf{MultiRoomN2S4} \\ \textbf{\tiny(Position)}}} & \textbf{Llama} & \textbf{0.9210} & \textbf{1842 / 2000} & \textbf{859 / 1000} & \textbf{0.26 $\pm$ 1.01} \\
		 & DeepSeek & 0.3310 & 662 / 2000 & 125 / 1000 & 2.91 $\pm$ 2.54 \\
		\ & Qwen & 0.7135 & 1427 / 2000 & 581 / 1000 & 1.12 $\pm$ 2.05 \\
		\hdashline[1pt/3pt]

		 \multirow{3}{*}{\shortstack[c]{\textbf{MultiRoomN2S4} \\ \textbf{\tiny(Representation)}}} & \textbf{Llama} & \textbf{0.9295} & \textbf{1859 / 2000} & \textbf{871 / 1000} & - \\
		 & DeepSeek & 0.2570 & 514 / 2000 & 21 / 1000 & - \\
		 & Qwen & 0.5585 & 1117 / 2000 & 286 / 1000 & - \\
		\hline
  
		 \multirow{3}{*}{\shortstack[c]{\textbf{MultiRoomN4S5} \\ \textbf{\tiny(Position)}}} & \textbf{Llama} & \textbf{0.5397} & \textbf{2159 / 4000} & \textbf{179 / 1000} & \textbf{3.69 $\pm$ 4.65} \\
		 & DeepSeek & 0.1200 & 480 / 4000 & 3 / 1000 & 7.92 $\pm$ 4.64 \\
		\ & Qwen & 0.3600 & 1440 / 4000 & 56 / 1000 & 5.23 $\pm$ 4.94 \\
		\hdashline[1pt/3pt]

		 \multirow{3}{*}{\shortstack[c]{\textbf{MultiRoomN4S5} \\ \textbf{\tiny(Representation)}}} & \textbf{Llama} & \textbf{0.5480} & \textbf{2192 / 4000} & \textbf{166 / 1000} & - \\
		 & DeepSeek & 0.1625 & 650 / 4000 & 7 / 1000 & - \\
		 & Qwen & 0.3517 & 1407 / 4000 & 38 / 1000 & - \\
		\hline
        \hline

		 \multirow{3}{*}{\shortstack[c]{\textbf{KeyCorridorS3R3} \\ \textbf{\tiny(Position)}}} & Llama & 0.9743 & 2923 / 3000 & \textbf{962 / 1000} & 0.18 $\pm$ 1.20 \\
		 & DeepSeek & 0.3337 & 1001 / 3000 & 18 / 1000 & 2.57 $\pm$ 2.54 \\
		\ & \textbf{Qwen} & \textbf{0.9810} & \textbf{2943 / 3000} & 958 / 1000 & \textbf{0.07 $\pm$ 0.63} \\
		\hdashline[1pt/3pt]

		 \multirow{3}{*}{\shortstack[c]{\textbf{KeyCorridorS3R3} \\ \textbf{\tiny(Representation)}}} & Llama & 0.9847 & 2954 / 3000 & 963 / 1000 & - \\
		 & DeepSeek & 0.9063 & 2719 / 3000 & 770 / 1000 & - \\
		 & \textbf{Qwen} & \textbf{0.9957} & \textbf{2987 / 3000} & \textbf{993 / 1000} & - \\
		\hline

		 \multirow{3}{*}{\shortstack[c]{\textbf{KeyCorridorS6R3} \\ \textbf{\tiny(Position)}}} & \textbf{Llama} & \textbf{0.9137} & \textbf{2741 / 3000} & \textbf{802 / 1000} & 1.47 $\pm$ 5.02 \\
		 & DeepSeek & 0.3650 & 1095 / 3000 & 44 / 1000 & 7.24 $\pm$ 7.47 \\
		\ & Qwen & 0.8810 & 2643 / 3000 & 732 / 1000 & \textbf{1.34 $\pm$ 4.25} \\
		\hdashline[1pt/3pt]

		 \multirow{3}{*}{\shortstack[c]{\textbf{KeyCorridorS6R3} \\ \textbf{\tiny(Representation)}}} & Llama & 0.9323 & 2797 / 3000 & 806 / 1000 & - \\
		 & DeepSeek & 0.7617 & 2285 / 3000 & 536 / 1000 & - \\
		 & \textbf{Qwen} & \textbf{0.9587} & \textbf{2876 / 3000} & \textbf{895 / 1000} & - \\
		\hline
        \hline

		 \multirow{3}{*}{\shortstack[c]{\textbf{ObstructedMaze1Dlh} \\ \textbf{\tiny(Position)}}} & \textbf{Llama} & \textbf{0.9870} & \textbf{2961 / 3000} & \textbf{984 / 1000} & \textbf{0.08 $\pm$ 0.82} \\
		 & DeepSeek & 0.4300 & 1290 / 3000 & 155 / 1000 & 2.98 $\pm$ 3.41 \\
		\ & Qwen & 0.6687 & 2006 / 3000 & 486 / 1000 & 2.18 $\pm$ 3.53 \\
		\hdashline[1pt/3pt]

		 \multirow{3}{*}{\shortstack[c]{\textbf{ObstructedMaze1Dlh} \\ \textbf{\tiny(Representation)}}} & \textbf{Llama} & \textbf{0.9877} & \textbf{2963 / 3000} & \textbf{986 / 1000} & - \\
		 & DeepSeek & 0.7390 & 2217 / 3000 & 452 / 1000 & - \\
		 & Qwen & 0.7417 & 2225 / 3000 & 561 / 1000 & - \\
		\hline

		 \multirow{3}{*}{\shortstack[c]{\textbf{ObstructedMaze1Dlhb} \\ \textbf{\tiny(Position)}}} & \textbf{Llama} & \textbf{0.4485} & \textbf{1794 / 4000} & 0 / 1000 & 4.33 $\pm$ 4.21 \\
		 & DeepSeek & 0.2288 & 915 / 4000 & \textbf{2 / 1000} & 4.49 $\pm$ 3.39 \\
		\ & Qwen & 0.4050 & 1620 / 4000 & 1 / 1000 & \textbf{3.85 $\pm$ 3.71} \\
		  \hdashline[1pt/3pt]

		 \multirow{3}{*}{\shortstack[c]{\textbf{ObstructedMaze1Dlhb} \\ \textbf{\tiny(Representation)}}} & Llama & 0.4858 & 1943 / 4000 & 0 / 1000 & - \\
		 & \textbf{DeepSeek} & \textbf{0.5182} & \textbf{2073 / 4000} & \textbf{14 / 1000} & - \\
		 & Qwen & 0.4820 & 1928 / 4000 & 1 / 1000 & - \\
		\hline
        \hline

	\end{tabular}}
	\end{adjustwidth}
\end{table*}

\end{document}